\documentclass{article}
\usepackage{spconf,amsmath,graphicx}
\usepackage{caption}
\DeclareCaptionFont{1pt}{\fontsize{9pt}{10pt}\selectfont}
\usepackage{tabularx}
\usepackage{booktabs}
\usepackage[]{hyperref}

\title{Unsupervised Single Image Underwater Depth Estimation}

\name{Honey Gupta, Kaushik Mitra
\thanks{This work was supported by Exploratory Research Grant (ELE/18-19/858/RFER/KAUS) from IIT Madras.}}
\address{Computational Imaging Lab, 
	Indian Institute of Technology Madras, India}
\begin{document}
\ninept
\maketitle
\begin{abstract}
Depth estimation from a single underwater image is one of the most challenging problems and is highly ill-posed. Due to the absence of large generalized underwater depth datasets and the difficulty in obtaining ground truth depth-maps, supervised learning techniques such as direct depth regression cannot be used. In this paper, we propose an unsupervised method for depth estimation from a single underwater image taken ``in the wild'' by using haze as a cue for depth. Our approach is based on indirect depth-map estimation where we learn the mapping functions between unpaired RGB-D terrestrial images and arbitrary underwater images to  estimate the required depth-map. We propose a method which is based on the principles of cycle-consistent learning and uses dense-block based auto-encoders as  generator  networks. We evaluate and compare our method both quantitatively and qualitatively on various underwater images with diverse attenuation and scattering conditions and show that our method produces state-of-the-art results for unsupervised depth estimation from a single underwater image. 

\end{abstract}
\section{Introduction}
\label{sec:intro}
\vspace{-0.1cm}
Exploration and mapping of underwater regions has always been a topic of interest. 
With the advancement in technology, it is now possible to visually explore the deep underwater regions that were unseen by humans few years back (like Mariana Trench). 
One of the most interesting fields of underwater exploration and mapping is 3D reconstruction,
% of sunken worlds; 
a key aspect of which is depth or 3D point estimation of the underwater scenes.

% Approaches for depth estimation of underwater regions either rely on specialized hardware sensors like time-of-flight (ToF)\cite{photon,photon2}, structured light \cite{structured} or LiDAR \cite{lidar2} or are based on image processing and estimate the depth map from single or set of images. Stereo based techniques\cite{stereo1,stereo2} have been proposed earlier but for feature matching the images have to be enhanced first, which in itself is a challenging problem. 

The current approaches for depth estimation of underwater regions either rely on specialized hardware sensors like time-of-flight (ToF)\cite{photon} and structured light\cite{structured} 
% or LiDAR \cite{lidar2} 
or use image processing methods to estimate the depth map from a single or a set of images. Stereo based techniques\cite{stereo1,stereo2} have also been used but they require feature matching, for which the images have to be enhanced first. This, in itself, is a challenging problem. 

\begin{figure}[!]
  \centering
  \includegraphics[width=\linewidth]{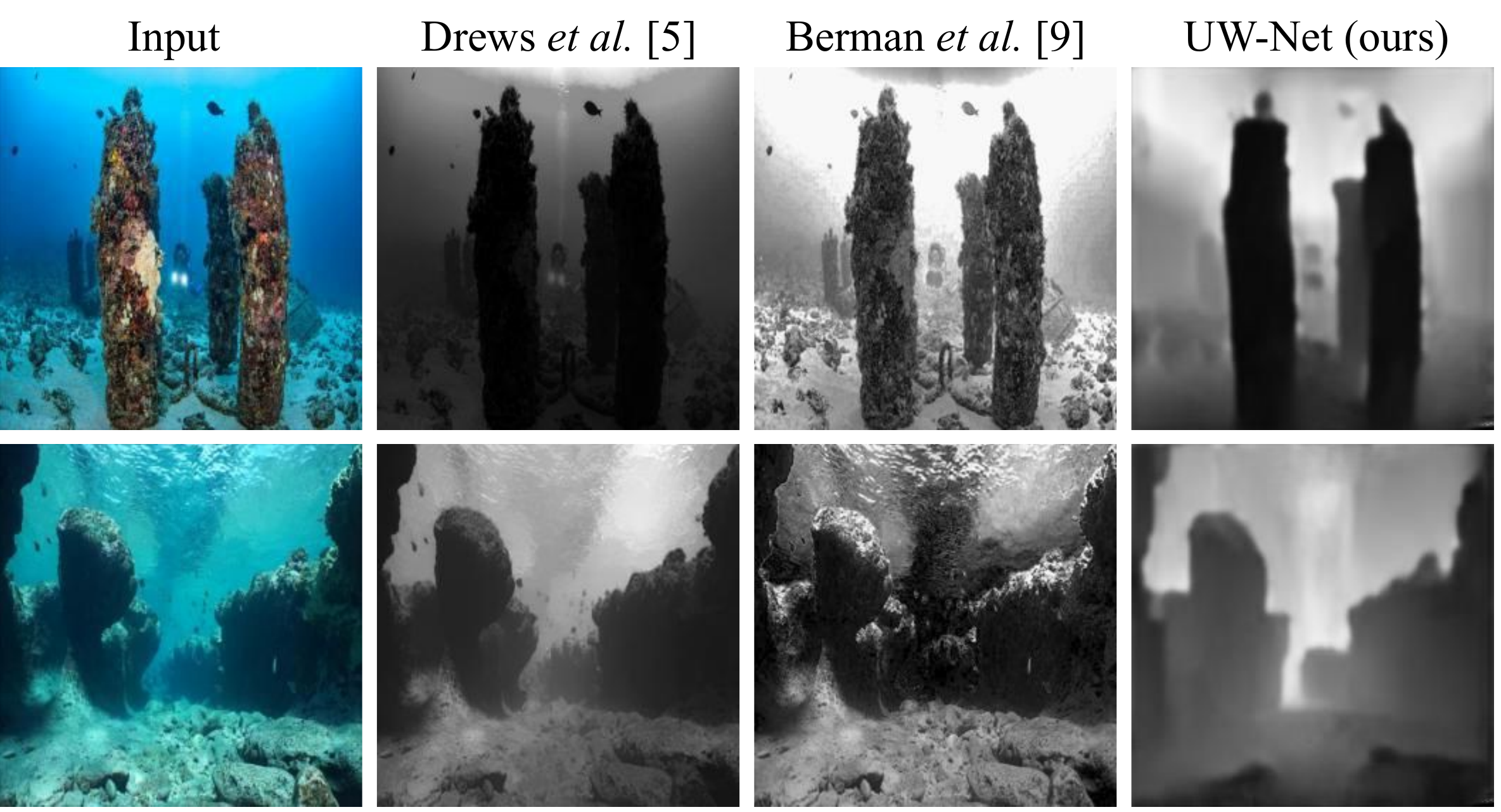} 
  
  \caption{
        % \fontsize{9pt}{0pt}
          \textit{Instead of following the usual approach of obtaining depth-map from transmission-map, we propose a novel depth estimation network based on unsupervised learning. Columns 2-3 show results from transmission-map based methods; column 4 shows our results.}}
          \label{fig: summary}
\end{figure}

Single underwater image depth estimation approaches \cite{udcp, ancuti1, peng, emberton, hazelines} mainly focus on image restoration.
They estimate the transmission-map as an intermediate step and derive depth-maps and restored images from that.
Transmission-map is related to the depth-map and scattering parameter via an exponential relationship $t(x) = e^{-\beta d(x)}$. 
Recovering the depth-map from the transmission-map implicitly assumes a specific scattering parameter, which might be incorrect. 
These approaches produce good results but due to the possibility of multiple depth-maps from a single transmission-map, a better approach would be to directly estimate depth from the underwater image.

Recently, deep learning based techniques have shown great promise for estimating depth from a single above-water image. Supervised approaches such as the methods proposed in  \cite{eigen} or \cite{kendall} use large above-water datasets to train deep neural networks. 
Unsupervised methods like  \cite{ mono} use stereo dataset for training.
% their network. 
However, these approaches are not suitable for estimating depth from underwater images due to the unavailability of large  datasets consisting of underwater (stereo) images and their corresponding 
ground truth depth-maps. To tackle this issue, we propose an unsupervised approach of transferring style between underwater and hazy above-water RGB-D images.  

In both underwater and hazy above-water images, the physical factors behind haze-like effect are different but the visual effects are similar i.e. visibility decreases as depth increases. The appearance of an underwater photo is governed by light scattering and attenuation as it propagates through water. This results in a depth-dependent function that relates the albedo of underwater objects to a hazy appearance from the camera's point of view. We aim to exploit this similarity between underwater and above-water hazy appearances to learn depth-map estimation. 

We take a generative approach of style-transfer and use haze as a cue to learn the mapping functions between hazy underwater and hazy above-water color image and depth-map. 
Our proposed network consists of two circularly connected dense-block \cite{densenet} based auto-encoders which learn to  estimate depth-map from a single underwater image based on the principle of cycle-consistent learning \cite{cyclegan}.  
Ideally, the transformed images should have the original structure of the source image and the desired style from the target image, but just having adversarial loss does not ensure that the original content and object structure of the images are preserved. To address this issue, we use SSIM \cite{ssim} as a structural loss  along with the adversarial loss for training the network. 
We also include a gradient sparsity loss term for the generated depth-map to reduce texture leakage artifacts in the depth-map. To summarize, the main contributions of this paper are as follows:

\begin{itemize}
    \item We propose an unsupervised deep-learning based method for estimating depth-map from a single underwater image. 
    \item We propose a network that contains circularly connected dense-block based auto-encoders that use structural and gradient sparsity loss to learn the mappings between hazy underwater and above-water hazy color image and depth-map. 
    \item We demonstrate that the proposed depth estimation method produces state-of-the-art results by comparing it with other unsupervised methods both quantitatively and qualitatively.
    % \item We also provide a comparison of different dense-prediction networks that depicts the effectiveness of these networks regarding depth estimation.
\end{itemize}

\section{Prior Work}

\subsection{Underwater depth estimation }
 Active underwater depth estimation methods have been proposed in the past few years like the ones by Maccarone \textit{et al.}\cite{photon} and Bruno \textit{et al.}\cite{structured}.
 %  McLeod \textit{et al.} \cite{lidar2} use LiDAR to get 3D. 
 The article by Campos \textit{et al.} \cite{sensors} gives an overview of different  sensors and methods that have been used for underwater 3D. 
 For single image depth estimation, the  dark channel prior (DCP) \cite{dcp} was one of the first methods to be used.
 Many variations of DCP have been introduced for underwater depth estimation and dehazing \cite{udcp,galdran,ancuti1}. Emberton \textit{et al.} \cite{emberton} segment regions of pure veiling-light and adapt the transmission-map to avoid artifacts. Peng  \textit{et  al.} \cite{peng}  estimate  the  scene  distance  via image blurriness. 
 Berman \textit{et al.} \cite{hazelines} take  into  account  multiple spectral profiles of different water types and choose the best result  based on color distribution. 
 Many of these methods focus on transmission-map estimation and obtain depth based on the relation between transmission-map and depth-map. Due to unknown scattering parameter and multiple possible solutions, these depth-map estimates are most likely to be incorrect.

\subsection{Depth estimation from single image}

Depth estimation approaches for single above-water images can be categorized as supervised and unsupervised methods. 
A variety of deep learning based supervised methods have been proposed, mostly comprising of training a model to predict depth from color image \cite{eigen,kendall,megadepth}. 
Other supervised techniques include joint estimation of depth with other related problems such as visual odometry \cite{demon}.  
Unsupervised methods for depth estimation include using stereo image pairs and photometric warp loss in place of reconstruction loss.
Godard \textit{et al.} \cite{mono} adopted this approach and used photometric loss between the warped left image and the right image.
These unsupervised methods require large stereo datasets such as KITTI for training and are unsuitable for underwater scenario.

\section{Underwater depth estimation Network: (UW-NET)}
\begin{figure}[t]
\centering
  \includegraphics[width=\linewidth]{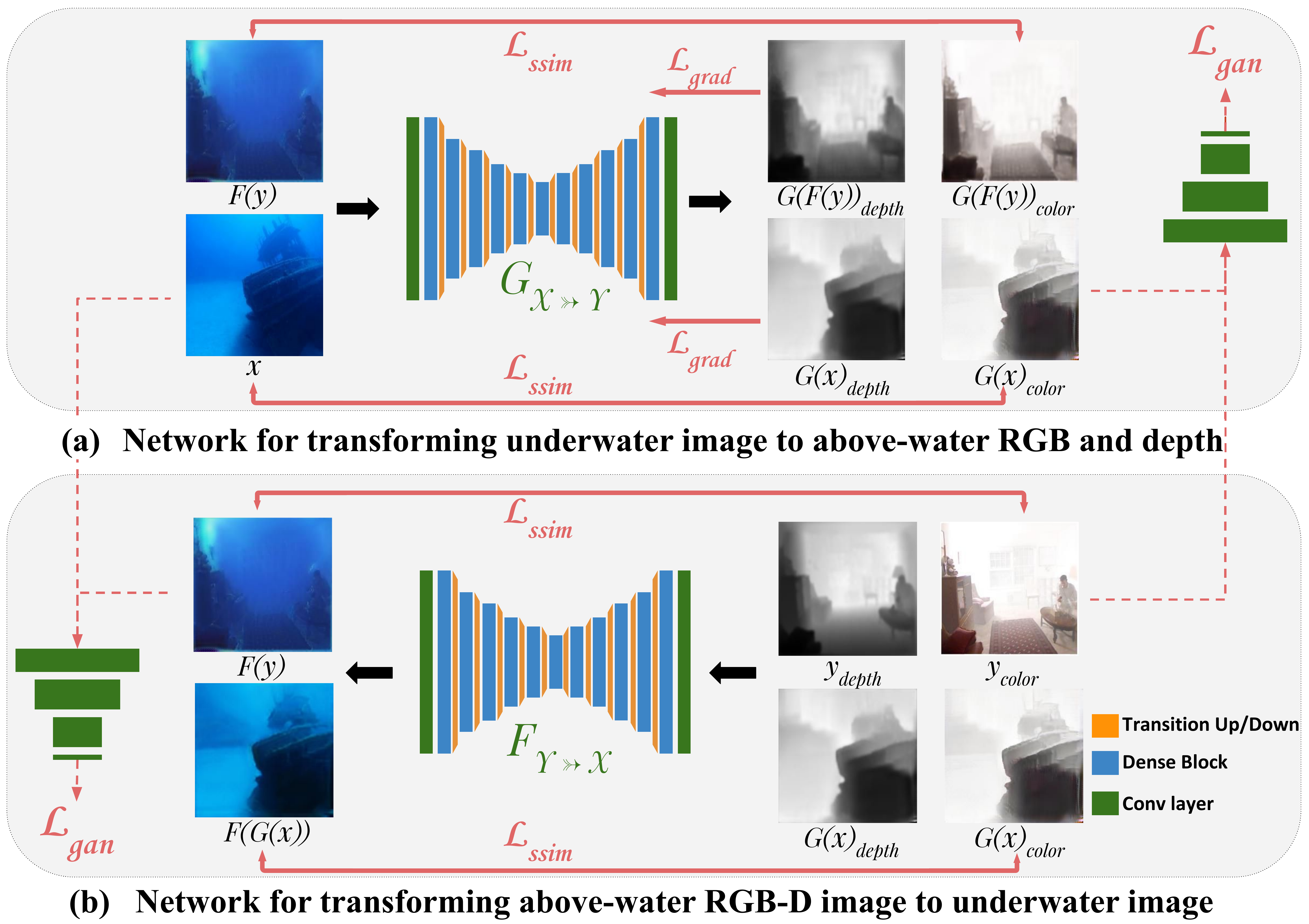}
    \mbox{} \hfill
  \caption{\label{fig: arch}%
         \textit{ Diagram representing the proposed UW-Net. 
         Generator $G$ transforms underwater images to above-water color and depth images whereas $F$ converts above-water color and depth to underwater color image. $G$ is constrained to learn depth estimation from underwater image due to the $\mathcal{L}_{cyc}$ between above-water $y$ and $G(F(y))$, thus producing depth-map from real underwater images.
        }} 

\end{figure}
\begin{figure*}[t]
  \centering
  \includegraphics[width=\linewidth]{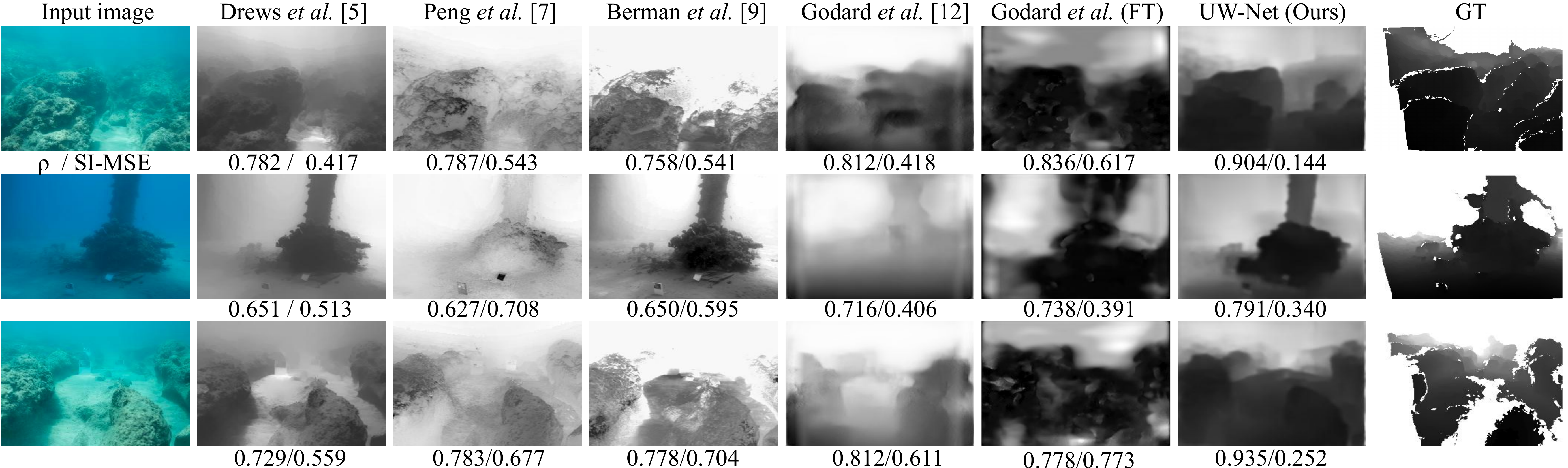}
  \caption{%
          Qualitative comparison of UW-Net with other underwater depth estimation methods and an unsupervised deep learning based method (Godard \textit{et al.} \cite{mono}) on few images from Berman \textit{et al.}'s \cite{hazelines} dataset.
          Our method produces meaningful depth-maps and has the highest Pearson coefficient and lowest scale-invariant error for the shown images. For further discussion on qualitative analysis, please refer to section \ref{sec:qual}. }
          \label{fig: comparison}
\end{figure*}

% \begin{table*}[t]
% \resizebox{\textwidth}{!}{
%  \centering
%   \begin{tabular}{|c|ccccccc|}
%   \hline
%     & DCP & Drews \textit{et al.}\cite{udcp} & Berman \textit{et al.}\cite{hazelines} & Peng \textit{et al.} \cite{peng}  & Godard \textit{et al.} \cite{mono} & Godard \textit{et al.} (FT) & UW-Net (Ours) \\
%     \hline
%     % & & & & & & & \\
    
% $\rho \uparrow$ 
% & 0.0869%195
% & 0.5560%728
% & 0.6698%699
% & 0.7221%059

% & 0.8039%855
% & 0.7321%573

% & \textbf{0.8289} \\ %361} \\

% SI-MSE $\downarrow$
% & 2.0223%588
% & 0.4503%395
% & 0.5186%241
% & 0.4853%221

% & 0.3115%540
% & 0.3637%359

% & \textbf{0.2211}\\%846}\\
% \hline
    
%   \end{tabular}}
%   \caption{\textit{Quantitative comparison of our method with state-of-the-art algorithms on Berman \textit{et al.}'s \cite{hazelines} dataset. The values were averaged over all the images. Higher $\rho$ values and lower SI-MSE values are better.}}
%   \label{table: comparison}
%  \end{table*}
 
\begin{table*}
 \begin{minipage}{\linewidth}
 \centering
 \resizebox{\linewidth}{!}{
  \begin{tabular}{cc|c|c|c|c|c|c}
  \toprule
    & DCP & Drews \textit{et al.}\cite{udcp} & Berman \textit{et al.}\cite{hazelines} & Peng \textit{et al.} \cite{peng}  & Godard \textit{et al.} \cite{mono} & Godard \textit{et al.} (FT) & UW-Net (Ours)\\
    \hline
    
$\rho \uparrow$ 
& 0.086%9195
& 0.556%0728
& 0.669%8699
& 0.722%1059

& 0.803%9855
& 0.732%1573

& \textbf{0.828} \\ %9361} \\

SI-MSE $\downarrow$
& 2.022%3588
& 0.450%3395
& 0.518%6241
& 0.485%3221

& 0.311%5%540
& 0.363%7%359

& \textbf{0.221}\\%1846}\\
\bottomrule
\end{tabular}}
\end{minipage}
\caption{\textit{Quantitative comparison of our method with state-of-the-art algorithms on Berman \textit{et al.}'s \cite{hazelines} dataset. The values were averaged over all the images. Higher $\rho$ values and lower SI-MSE values are better.}} \label{table: comparison}
 \end{table*}
 
In our proposed method, we learn to transform images from a domain of hazy underwater images $X$ to a domain of hazy terrestrial RGB-D images $Y$ and vice versa.
The network learns the mapping between these domains with the help of unpaired images belonging to both the image distributions $x \sim p_{uw}(x)$ and $y \sim p_{aerial}(y)$. 
% Please refer Fig.\ref{fig: arch} for a visual representation of our method. 
The network consists of two generator auto-encoder networks which learn the two mapping functions $G : X  \rightarrow  Y $ and $F : Y  \rightarrow  X$.  
It also consists of two discriminators $D_{X}$ and $D_{Y}$. 
The generator and the discriminator learn through adversarial learning \cite{gan}. The discriminator $D_{X}$ tries to classify its input as real ($\{x\}$) or fake ($\{F(y)\}$) underwater image .
The discriminator $D_{Y}$ similarly classifies above-water images as real or fake. 
% tries to classify its input as real above-water image ($\{y\}$) or fake ($\{G(x)\}$).
The generators' objective function is a weighted combination of four loss terms: 
\noindent
\begin{equation} \label{totalloss}
\begin{split}
& \mathcal{L}_{total}(G, F, D_{X}, D_{Y}) = \hfill \\
& \mathcal{L}_{cyc} + \gamma_{gan}\mathcal{L}_{gan} + \gamma_{ssim}\mathcal{L}_{ssim} + \gamma_{grad}\mathcal{L}_{grad} \hfill
\end{split}
\end{equation}\label{eq:total}

\noindent The first two terms are cyclic $\mathcal{L}_{cyc}$ and adversarial loss $\mathcal{L}_{gan}$, similar to CycleGAN\cite{cyclegan}. We use $\mathcal{L}_{1}$ for $\mathcal{L}_{cyc}$ loss and since the convergence of least squares version of $\mathcal{L}_{gan}$ is better as compared to the log-likelihood loss \cite{l2}, we use the least squares version of $\mathcal{L}_{gan}$.

\vspace{0.15cm}
\noindent \begin{equation}\label{eq1}
\begin{split}
\mathcal{L}_{gan}(G, D_{Y}, X, Y) & = \mathbf{E}_{y \sim p_{aerial}(y)} [(D_{Y}(y) - 1)^{2}]  \\ 
& + \mathbf{E}_{x \sim p_{uw}(x)}[D_{Y}(G(x))^{2}]   
\end{split}
\end{equation}
% \vspace{0.05cm}

Adversarial loss is the only loss term that influences the image quality of the generated image after half a cycle in the CycleGAN\cite{cyclegan} method, which does not guarantee preservation of structural properties of the input image in the transformed image.
In order to alleviate this problem, we introduce structural similarity loss $\mathcal{L}_{ssim}$  in the objective function.
It is defined as $1-SSIM(\{,\})$ calculated between $\{x, G(x) \}$, $\{y, F(y) \}$, $\{ G(x), F(G(x)) \}$ and $\{F(y), G(F((y)) \}$. 
In case of $y$, $G(x)$ and $G(F(y))$, we consider only the RGB channels, as including $\mathcal{L}_{ssim}$ for depth introduces unwanted artifacts. 
In our training dataset, above-water RGB-D images belong to indoor scenes and this creates a mismatch between the geometry of objects in the above-water and underwater images. Due to this undesired textures get introduced in the estimated depth-maps. We introduce a sparsity of gradient loss ${L}_{grad}$ to reduce this texture. It is defined as
\begin{equation}
\begin{split}
\mathcal{L}_{grad} = \sum \; (\| \nabla_{x} \ d \|_{1} + \| \nabla_{y} \ d \|_{1})
\end{split}
\end{equation}
\noindent
where $d$ is the depth channel of $G(x)$. 
\\\\
\noindent \textbf{Network architecture}\label{sec:arch}. 
Fig.\ref{fig: arch} depicts a schematic diagram of our network architecture. 
% As suggested by Zhu \textit{et al.} \cite{cyclegan}, the model is similar to training two autoencoders jointly, $G \circ F: X \rightarrow X$ and $F \circ G: Y \rightarrow Y$. 
We use a dense-block based auto-encoder network \cite{tiramisu} for the generator. It consists of 5 dense blocks (DB) + transition down/up modules, where each DB has 5 layers with 16 filters.
For discriminator, we use 70x70 PatchGANs, similar to \cite{cyclegan}. 

Our proposed method is based on indirect learning of the depth map. 
An above-water image sample, $y \in Y$, is a four-channel data sample $y = [y_{color}, y_{depth}]$, where the fourth channel is the depth map. 
The output of our generator $G$ is color as well as depth and so is the input of generator $F$.
The discriminator $D_Y$ also takes both color and depth images as input. We observed that including depth in discriminator $D_Y$ helps the network produce better depth-maps. The discriminator $D_X$ takes only color images as input. 
A 4-channel image $y$ is passed through $F$ to convert it to an underwater image. 
% After transformation by the mapping function $F$,
The obtained underwater image $F(y)$ is passed through $G$ to obtainobtain $G(F(y))$, which completes one cycle.
Similarly, we transform $x \in X \rightarrow G(x) \rightarrow F(G(x))$, which forms another cycle. 
$D_X$ enforces $F(y)$ to be close to underwater image-set $X$. 
Due to cyclic loss and ``realness" of fake underwater image, the network $G$ learns to estimate depth from single underwater image. 
% The discriminator $D_{x}$ tries to classify image $x$ as real and $F(y)$ as fake. Similarly, discriminator $D_{y}$ tries to classify $y$ as real and $G(x)$ as fake. 

\section{Experiments}\label{sec:exp}

\subsection{Training details} \label{dataset}
For our experiments\footnote{Please refer our project page: http://bit.ly/uw-net for code and results.}, we collected $1343$ real underwater images from different sources on the internet. We focused on collecting underwater images with different illumination and scattering conditions. 
% We tried to compile a diverse set of images, which subsequently creates a more generalized model. 
For hazy above-water images, we have used the D-Hazy dataset \cite{dhazy}. The dataset consists of $1449$ synthetic hazy indoor images. To reduce sensor noise in the ground truth depth-maps, 
% the depth map has a lot of texture, which was resulting in artifacts in our estimated depth maps. Therefore 
we performed bilateral filtering on the depth images before training. 
All the images were down-sampled to a size of  $256 \times 256$ and trained on random $128 \times 128$ patches for 400 epochs using ADAM optimizer, with a learning rate of $1 \times 10^{-4}$.
While testing, we took complete images of size $256 \times 256$.  The experiments were performed on Nvidia GTX 1080Ti, using TensorFlow framework. We experimented with the weights for each of the loss term and found the optimum values for $\gamma_{gan}, \ \gamma_{ssim} \ \mathrm{and} \ \gamma_{grad}$ to be 5, 1 and 1 respectively. 

\begin{figure*}[t]
    \centering
    \includegraphics[width=\linewidth]
    {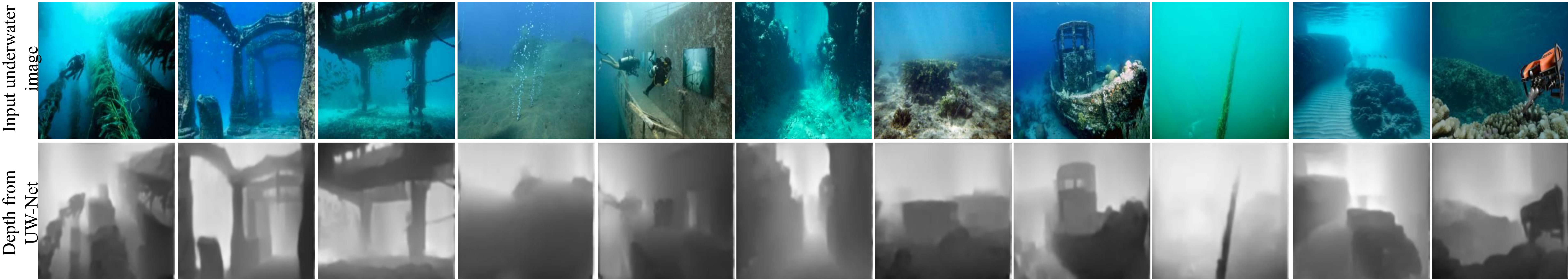}
    
    \caption{
        \textit{\textbf{UW-Net} depth-map estimates for randomly downloaded underwater images from internet belonging to different illumination and scattering conditions. Comparison with other methods can be found at      \href{http://bit.ly/2JkXZ0B}{https://goo.gl/tB22gU}.}} 
        \label{fig:more_images}
\end{figure*}

\vspace{-0.15cm}
\subsection{Comparison with other methods}
We compare our depth estimation method with other underwater depth estimation methods by Drews \textit{et al.} (UDCP) \cite{udcp}, Peng \textit{et al.} \cite{peng} and Berman \textit{et al.} \cite{hazelines}. 
We also compare with a deep learning based depth estimation method by Godard \textit{et al.} \cite{mono} which uses stereo image pairs and photometric warp loss for training. 
We used the codes publicly released by Drews \textit{et al.}, Berman \textit{et al.} and Godard \textit{et al.} for evaluation. For Peng \textit{et al.}, we used the code provided by the authors.
Default hyper-parameters were used for all the methods during evaluation.
For all the transmission-map based methods, we took negative-log to obtain the depth map. 
For Godard \textit{et al.}\cite{mono}, we show two results, one with their pre-trained KITTI model and the other with a fine-tuned (FT) underwater model (Godard \textit{et al.}(FT)). 
% We tried both KITTI and Cityscapes model but KITTI results were better. 
For the fine-tuned model, we re-trained the KITTI model for $20$ epochs on $2000$ stereo pairs randomly picked from three different subsets of CADDY dataset \cite{caddy}.
We fine-tuned our trained model for $20$ epochs on Berman \textit{et al.}'s dataset for better results.

\vspace{-0.4cm}
\subsubsection{Quantitative analysis}\label{sec:quan}
% Metric definition
For the quantitative comparison, we have used the dataset released by Berman \textit{et al.}\cite{hazelines}. It contains $114$ underwater RGB images collected from 4 different locations along with the ground-truth depth-map(GT).
We have used $98$ images belonging to regions: Katzaa, Nachsholim and Satil. We did not use images from the region Michmoret as they have very less natural objects. 
Only pixels with a defined depth-value in GT were used for calculation. 
Since ours as well as all other methods produce depth-maps up to scale, we use two scale-invariant metrics for comparison: log Scale-Invariant Mean Squared Error (SI-MSE), as proposed by Eigen \textit{et al.} in \cite{eigen} and Pearson correlation coefficient ($\rho$) defined as $\rho_{X,Y} = \frac{cov(X,Y)}{\sigma_{x}\sigma_{y}}$. 

Table \ref{table: comparison} shows the quantitative results. We observe that our method performs better than others and produces the highest Pearson coefficient and the least scale-invariant error. 
The closest to our approach is the pre-trained Godard \textit{et al.} model, as it might be a somewhat generalized model. 
We think that fine-tuning reduces the accuracy as it over-fits the model for specific underwater conditions and hence fails for images belonging to other regions. 
Among other underwater depth-estimation methods, Peng \textit{et al.} perform the best.

\begin{figure}[t]
    \centering
    \includegraphics[width=0.99\linewidth]{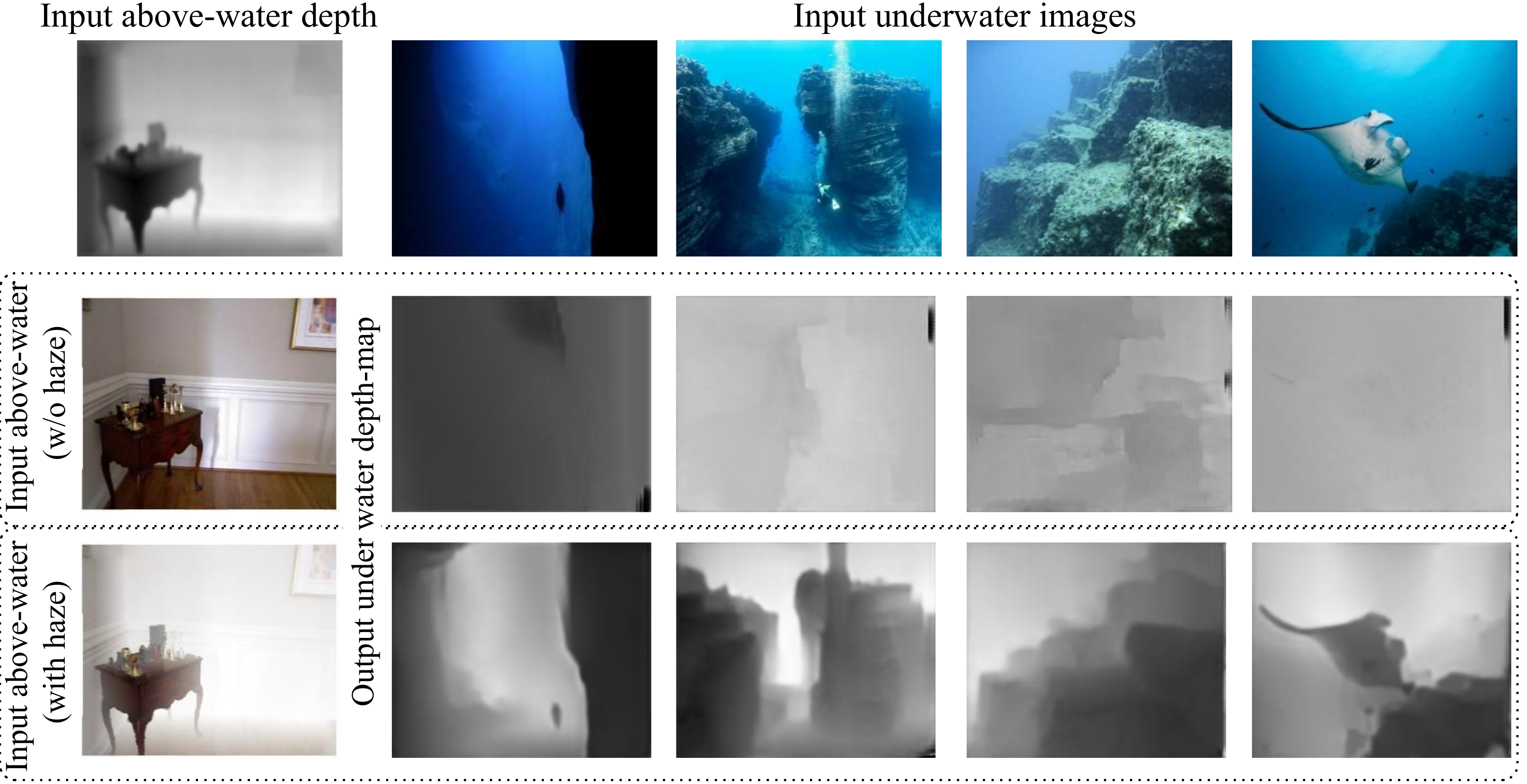}
    \caption{
        \textit{With clear RGB above-water images during training, the network fails to learn the mapping between underwater image and the depth-map (row 2) but with hazy above-water images, the network is able to learn the mapping quite well (row 3).}}
        \label{fig:rgbvsrgbd}
\end{figure}

\vspace{-0.1cm}
\subsubsection{Qualitative analysis}\label{sec:qual}

Fig.\ref{fig: comparison} depicts a comparison of our results with other methods for few single underwater images taken from Berman \textit{et al.}'s dataset. Results for collected underwater images from internet with different illumination and scattering conditions can be found in Fig.\ref{fig: summary}, Fig.\ref{fig:more_images} and at this link:  
\begingroup
    \fontsize{9pt}{0pt}\selectfont
        \href{http://bit.ly/2Hs5fnO}{\textit{https://goo.gl/ASJv2f}}. 
\endgroup

In Fig.\ref{fig: comparison}, column 1 shows the input underwater images, columns 2, 3 and 4 show the depth from image-processing based methods and columns 5 and 6 show results by Godard \textit{et al.} w/o and with fine-tuning. 
In columns 2-4,
% it might looks like it is capturing depth well, but if 
a closer look at the sea-floor area near to the camera reveals that these methods fail to capture relative depth of the scene with respect to the camera. Moreover, there is excessive texture leakage in the estimated depth-maps as these methods mainly capture the transmission properties of the scene. 
For Godard \textit{et al.}, we see improper results due to the mismatch between the train-test data samples. Even for the fine-tuned model, 
% since CADDY dataset is not a generalized dataset,
the results worsen as the images belong to different underwater regions. Due to unavailability of a diverse stereo dataset, stereo based methods are bound to perform moderately. 
On the other hand, our depth-maps seem much more reasonable with a linear depth variation and have higher $\rho$ and lower scale-invariant error. 
Even for varying illumination, we observe that our network successfully captures the depth information. 

% \begin{figure}[t]
%     \centering
%     \includegraphics[width=\linewidth]
%     {images/Results_2.pdf}
    
%     \caption{
%         \textit{Our depth-map estimates for randomly chosen underwater images from our collected dataset with different illumination and scattering conditions}}
%         \label{fig:more_images}
% \end{figure}
\begin{table}[h!]
    \centering
        \resizebox{\linewidth}{!}{\begin{tabular}{|c|cccccc|rlrr}
        \hline
             & EDSR & Hourglass & U-Net & ResNet & DenseRes & DenseNet \\
        \hline
$\rho \uparrow$

& 0.732
& 0.758
& 0.769
& 0.773
& 0.768
& \textbf{0.829} \\

SI-MSE $\downarrow$

& 0.364 
& 0.310 
& 0.279
& 0.297
& 0.271
& \textbf{0.221} \\
\hline
\end{tabular}}
\caption{\textit{Quantitative comparison of different auto-encoder networks as generator network on Berman \textit{et al.}'s \cite{hazelines} dataset.}}
\label{table:generator}
\end{table}

\vspace{-0.2cm}
\subsection{Ablation Study}

\subsubsection{Depth from clean vs. hazy RGB-D}\label{rgbdvsrgb}

Here we show the importance of using hazy above-water images during training and validate that our network learns to capture depth by using haze as a cue. 
We trained a separate network with the same configuration as UW-Net, but in place of D-Hazy dataset, we used the original NYU dataset (without haze).
% .we use In Fig. \ref{fig: rgbvsrgbd}, column 2 shows a sample input above-water image and the corresponding underwater depth map obtained. 
% we show the result for one sample of our unsupervised depth estimation method when clear color aerial images were used for training the network and in part (b) we show the estimated depth for the same sample when hazy color aerial images were used for training the network. We used the same underwater dataset and same input depth map for both the networks. For the first network, we used images from D-Hazy \cite{dhazy} dataset (which is derived from NYU\cite{nyu} dataset) and for the second, we directly use the NYU dataset. 
In Fig.\ref{fig:rgbvsrgbd} row 2, we observe that the network using clear above-water images produces poor underwater depth-map estimates, whereas in row 3, UW-Net produces quite good depth-maps. 
We think this is because of the simpler relation between depth and haze as compared to depth and object geometry. 
Moreover, using hazy above-water images gives rise to depth-dependent attenuation in the transformed $F(y)$, which in turn helps $G$ to learn the relation between depth and attenuation. 

\vspace{-0.1cm}
\subsubsection{Choice of Generator Network}

We experimented with different auto-encoder networks to find the best network architecture for the generator. The originally proposed Cycle-GAN \cite{cyclegan} has ResNet \cite{resnet} based generator network, which performs quite well. Table \ref{table:generator} shows the Pearson coefficient and scale-invariant error of the results obtained by training our network with U-Net \cite{unet}, EDSR(w/o upsampling) \cite{edsr}, Hourglass \cite{hourglass}, ResNet \cite{resnet}, FC-DenseNet56 \cite{tiramisu} followed by 3 residual blocks (DenseRes) and DenseNet \cite{tiramisu} models. Please refer to the referenced papers for further details regarding the networks.
It can be seen from Table \ref{table:generator} that DenseNet produces the least scale-invariant error and the highest Pearson correlation coefficient, making it the best choice. Closest to DenseNet are ResNet \cite{resnet} and DenseRes architectures. 

% \vspace{-0.2cm}
\section{Conclusion}
We proposed an unsupervised method that uses haze as a cue to estimate depth-map from a single underwater image. 
Our method is based on cycle-consistent learning and uses dense-block based auto-encoder for generator networks. 
The proposed network learns indirectly to estimate the depth-map and uses SSIM and gradient sparsity loss for better estimation. 
We compared our method with existing state-of-the-art single image underwater depth estimation methods and showed that our method 
performs better than all other methods both qualitatively and quantitatively. 
% in terms of Pearson coefficient and scale-invariant error.

% \begin{figure}[htb]

% \begin{minipage}[b]{1.0\linewidth}
%   \centering
%   \centerline{\includegraphics[width=8.5cm]{image1}}
% %  \vspace{2.0cm}
%   \centerline{(a) Result 1}\medskip
% \end{minipage}
% %
% \begin{minipage}[b]{.48\linewidth}
%   \centering
%   \centerline{\includegraphics[width=4.0cm]{image3}}
% %  \vspace{1.5cm}
%   \centerline{(b) Results 3}\medskip
% \end{minipage}
% \hfill
% \begin{minipage}[b]{0.48\linewidth}
%   \centering
%   \centerline{\includegraphics[width=4.0cm]{image4}}
% %  \vspace{1.5cm}
%   \centerline{(c) Result 4}\medskip
% \end{minipage}
% %
% \caption{Example of placing a figure with experimental results.}
% \label{fig:res}
% %
% \end{figure}

%\vfill
%\pagebreak

% \section{COPYRIGHT FORMS}
% \label{sec:copyright}

% You must include your fully completed, signed IEEE copyright release form when
% form when you submit your paper. We {\bf must} have this form before your paper
% can be published in the proceedings.

\bibliographystyle{IEEEbib}
\bibliography{refs}

\end{document}